\documentclass{article}

\PassOptionsToPackage{numbers, compress}{natbib}

\usepackage[preprint]{neurips_2023}

\usepackage[utf8]{inputenc} %
\usepackage[T1]{fontenc}    %
\usepackage[colorlinks]{hyperref}       %
\usepackage{url}            %
\usepackage{booktabs}       %
\usepackage{amsfonts}       %
\usepackage{nicefrac}       %
\usepackage{microtype}      %
\usepackage{xcolor}         %
\usepackage{graphicx}
\usepackage{amsmath}
\usepackage{enumitem}

\title{Diffusion Models Beat GANs on Image Classification}

 \author{%
   Soumik Mukhopadhyay$^*$ \And Matthew Gwilliam$^*$ \And Vatsal Agarwal \And
   Namitha Padmanabhan \And Archana Swaminathan \And Srinidhi Hegde \And
   Tianyi Zhou \And Abhinav Shrivastava \\
   \\
   University of Maryland, College Park\\
}

\begin{document}

\maketitle

\def\thefootnote{*}\footnotetext{These authors contributed equally to this work}

\begin{abstract}
  While many unsupervised learning models focus on one family of tasks, either generative or discriminative, we explore the possibility of a unified representation learner: a model which uses a single pre-training stage to address both families of tasks simultaneously. 
  We identify diffusion models as a prime candidate. 
  Diffusion models have risen to prominence as a state-of-the-art method for image generation, denoising, inpainting, super-resolution, manipulation, etc. 
  Such models involve training a U-Net to iteratively predict and remove noise, and the resulting model can synthesize high fidelity, diverse, novel images. 
  The U-Net architecture, as a convolution-based architecture, generates a diverse set of feature representations in the form of intermediate feature maps. 
  We present our findings that these embeddings are useful beyond the noise prediction task, as they contain discriminative information and can also be leveraged for classification. 
  We explore optimal methods for extracting and using these embeddings for classification tasks, demonstrating promising results on the ImageNet classification task.
  We find that with careful feature selection and pooling, diffusion models outperform comparable generative-discriminative methods such as BigBiGAN for classification tasks.
  We investigate diffusion models in the transfer learning regime, examining their performance on several fine-grained visual classification datasets. 
  We compare these embeddings to those generated by competing architectures and pre-trainings for classification tasks.
\end{abstract}

\section{Introduction}
\label{sec:introduction}

Unified unsupervised image representation learning is a critical but challenging problem.
Many computer vision tasks can be broadly classified into two families, discriminative and generative.
With discriminative representation learning, one seeks to train a model that can apply labels to images or parts of images. 
For generative learning, one would design a model that generates or edits images, and performs similar tasks like inpainting, super-resolution, etc.
Unified representation learners seek to achieve both simultaneously, and the resulting model would be able to both discriminate and generate novel visual artifacts.

\begin{figure}[h!]
    \centering
    \includegraphics[width=1.0\linewidth]{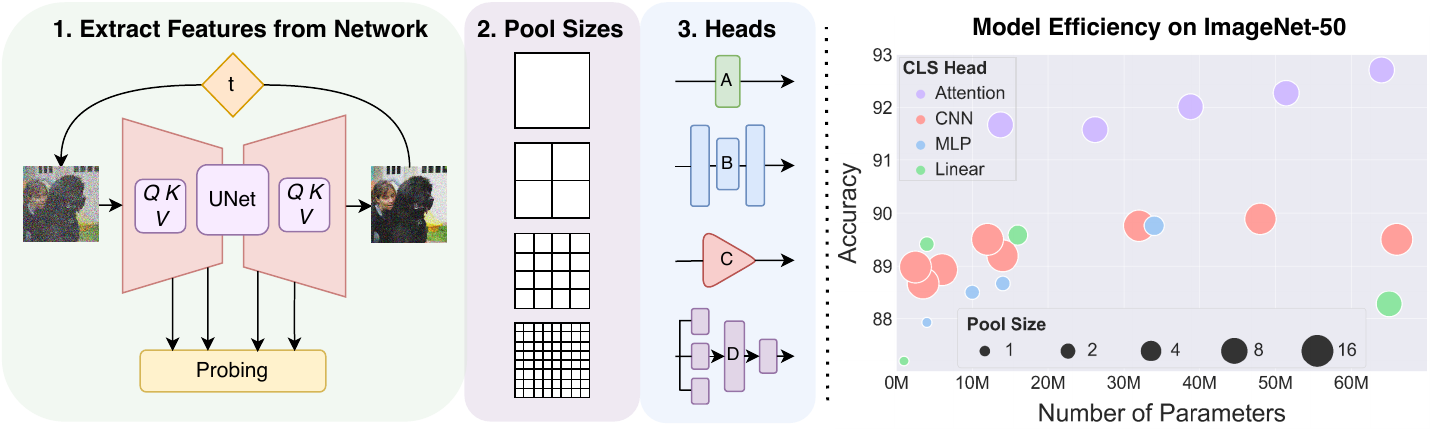}
    \caption{An overview of our method and results. We propose that diffusion models are unified self-supervised image representation learners, with impressive performance not only for generation, but also for classification. We explore the feature extraction process in terms of U-Net block number and diffusion noise time step. We also explore different sizes for the feature map pooling. We examine several lightweight architectures for feature classification, including linear (A), multi-layer perceptron (B), CNN (C), and attention-based heads (D). We show the results on such explorations on the right, for classification heads trained on frozen features for ImageNet-50~\cite{vangansbeke2020scan}, computed at block number 24 and noise time step 90. See Section~\ref{subsec:main_results} for a detailed discussion.}
    \label{fig:teaser_fig}
\end{figure}

Such unified representation learning is an arduous undertaking.
BigBiGAN~\cite{donahue2016adversarial,donahue2019large} is one of the earliest deep learning methods to address both families of tasks simultaneously.
However, more recent approaches outperform BigBiGAN in terms of both classification and generation performance by more specialized models.
Beyond BigBiGAN's key accuracy and FID deficiencies, it is also much more burdensome to train than other methods; its encoder makes it larger and slower than comparable GANs, and its GAN makes it more expensive than ResNet-based discriminative methods~\cite{he2015deep}.
PatchVAE~\cite{Gupta_2020_CVPR} attempts to adapt VAE~\cite{kingma2022autoencoding} to perform better for recognition tasks by focusing on learning mid-level patches.
Unfortunately, its classification gains still fall well short of supervised methods, and come at great cost to image generation performance.
Recent works have taken valuable steps by delivering good performance in both generation and classification, both with~\cite{chang2022maskgit} and without~\cite{li2022mage} supervision.
However, this field is relatively underexplored in comparison to the volume of work in self-supervised image representation learning, and therefore, unified self-supervised representation learning remains largely under-addressed.

As a result of previous shortcomings, some researchers have argued that there are inherent differences between discriminative and generative models, and the representations learned by one are not well-suited for the other~\cite{DBLP:journals/corr/abs-2006-07733}.
Generative models naturally need representations that capture low-level, pixel and texture details which are necessary for high fidelity reconstruction and generation.
Discriminative models, on the other hand, primarily rely on high-level information which differentiates objects at a coarse level based not on individual pixel values, but rather on the semantics of the content of the image.
Despite these preconceptions, we suggest that the early success of BigBiGAN is endorsed by recent approaches such as MAE~\cite{DBLP:journals/corr/abs-2111-06377} and MAGE~\cite{li2022mage}, where the model must tend to low-level pixel information, but learns models which are also very good for classification tasks.

Furthermore, state-of-the-art diffusion models have already achieved great success for generative objectives. 
However, their classification capacity is largely ignored and unexplored.
Thus, rather than build a unified representation learner from the ground up, we posit that state-of-the-art diffusion models, which are powerful image generation models, already possess potent emergent classification properties.
We demonstrate their impressive performance on these two very different tasks in Figure~\ref{fig:teaser_fig}.
Our method for utilizing diffusion models results in much better image generation performance than BigBiGAN, with better image classification performance as well.
Thus, in terms of optimizing for both classification and generation simultaneously, we show that diffusion models are already near state-of-the-art unified self-supervised representation learners.

One of the main challenges with diffusion models is feature selection. 
In particular, the selection of noise steps and feature block is not trivial. 
So, we investigate and compare the suitability of the various features.
Additionally, these feature maps can be quite large, in terms of both spatial resolution and channel depth.
To address this, we also suggest various classification heads to take the place of the linear classification layer, which can improve classification results, without any addition of parameters or sacrifice in generation performance.
Critically, we demonstrate that with proper feature extraction, diffusion models work very well as classifiers out-of-the-box, such that diffusion models can be used for classification tasks without the need to modify the diffusion pre-training.
As such, our approach is flexible for any pre-trained diffusion model and can thus benefit from future improvements to such models in terms of size, speed, and image quality.

We also investigate the performance of diffusion features for transfer learning on downstream tasks, and we compare the features themselves directly to those from other methods. 
For downstream tasks, we choose fine-grained visual classification (FGVC), an appealing area to use unsupervised features due to implied scarcity of data for many FGVC datasets.
This task is of particular interest with a diffusion-based method since it does not rely on the sorts of color invariances that previous works suggest may limit unsupervised methods in the FGVC transfer setting~\cite{xiao2021contrastive,Gwilliam_2022_CVPR}.
To compare the features, we rely on the popular centered kernel alignment (CKA)~\cite{pmlr-v97-kornblith19a}, which allows for a rich exploration of the importance of feature selection as well as how similar diffusion model features are to those from ResNets~\cite{he2015deep} and ViTs~\cite{dosovitskiy2020image}.

In summary, our contributions are as follows:
\begin{itemize}
    \item We demonstrate that diffusion models can be used as \textit{unified} representation learners, with 26.21 FID (-12.37 \textit{vs.} BigBiGAN) for unconditional image generation and 61.95\% accuracy (+1.15\% \textit{vs.} BigBiGAN) for linear probing on ImageNet.
    \item We present analysis and distill principles for extracting useful feature representations from the diffusion process.
    \item We compare standard linear probing to specialized MLP, CNN, and attention-based heads for leveraging diffusion representations in a classification paradigm.
    \item We analyze the transfer learning properties of diffusion models, with fine-grained visual categorization (FGVC) as a downstream task, on several popular datasets.
    \item We use CKA to compare the various representations learned by diffusion models, both in terms of different layers and diffusion properties, as well as to other architectures and pre-training methods.
\end{itemize}

\section{Related Work}
\label{sec:related_work}

\paragraph{Generative Models.}

Generative Adversarial Networks (GANs)~\cite{goodfellow2020generative} constitute a class of deep neural networks which are capable of generating novel images in a data distribution given a random latent vector $z\in\mathcal{Z}$ as input, and are trained by optimizing a min-max game objective. 
GANs can be class-conditioned, where they generate images given noise and class input, or unconditional, where they generate random images from noise alone.
Popular examples of GANs which have produced high quality images include PGGAN~\cite{karras2017progressive}, BigGAN~\cite{brock2018large}, and StyleGANs~\cite{Karras_2019_CVPR,karras2020analyzing,karras2020training,karras2021alias,Sauer2021ARXIV}.
Recent work in GAN inversion finds that images can be mapped to GAN latent space~\cite{DBLP:journals/corr/abs-2101-05278}, meaning that the GAN learns a representation for the image in noise/latent space. 
Some of these approaches directly optimize latent vectors to reconstruct the input image~\cite{DBLP:journals/corr/abs-1904-03189}.
Others train encoders to generate the latent vector corresponding to a given input image~\cite{DBLP:journals/corr/abs-2008-00951,DBLP:journals/corr/abs-2102-02766}.
Hybrid approaches are also popular, where an encoder generates a latent vector which is then optimized to generate better reconstructions~\cite{zhu2016generative,zhu2020domain,roich2022pivotal,alaluf2022hyperstyle}.

Diffusion denoising probabilistic models (DDPM)~\cite{ho2020denoising}, a.k.a.\ diffusion models, are a class of likelihood-based generative models which learn a denoising Markov chain using variational inference. Diffusion models have proven to produce high-quality images~\cite{dhariwal2021diffusion} beating previous SOTA generative models like BigGAN~\cite{brock2018large}, VQVAE-2~\cite{razavi2019generating} on FID metric on ImageNet\cite{deng2009imagenet}. These models enjoy the benefit of having a likelihood-based objective like VAEs as well as high visual sample quality like GANs even on high variability datasets. Recent advances in this area have also shown amazing results in text-to-image generation including works like DALLE2~\cite{ramesh2022hierarchical}, Imagen~\cite{saharia2022photorealistic}, and Stable Diffusion~\cite{rombach2022high}.  Application of these models is not just limited to generation but spans tasks like object detection~\cite{chen2022diffusiondet},  and image segmentation~\cite{burgert2022peekaboo}.
While these are all trained and evaluated for generative tasks, we observe they have discriminative capacity as well, and thus investigate their potential for classification tasks.

\paragraph{Discriminative Models.} 
Discriminative models learn to represent images, and extract useful information from images that can then be used to solve downstream tasks. 
Early representation learning methods tried training neural network backbones with partially degraded inputs and learn image representation by making the model predict the rest of the information in the actual image like Colorisation~\cite{zhang2016colorful}, Jigsaw~\cite{noroozi2016unsupervised}, PIRL~\cite{misra2020self}, Inpainting~\cite{pathak2016context}.
More recently, many approaches have emerged that revolve around a contrastive loss objective, maximizing distance between positive-negative pairs, such as SimCLR~\cite{chen2020simple,chen2020big}, MoCo~\cite{DBLP:journals/corr/abs-2003-04297,chen2020improved,chen2021mocov3}, Barlow Twins~\cite{pmlr-v139-zbontar21a}, and ReLICv2~\cite{tomasev2022pushing}.
On the other hand, BYOL~\cite{DBLP:journals/corr/abs-2006-07733}, SimSiam~\cite{chen2021exploring}, and VICReg~\cite{Bardes2021VICRegVR} introduce methods that work without negative samples.
DeepCluster~\cite{caron2018deep} uses offline clustering whereas SwAV~\cite{caron2020unsupervised} introduces online clustering and multi-view augmentation methods to get a better representation.
DINO~\cite{caron2021emerging} uses self supervised knowledge distillation between various views of an image in Visual Transformers~\cite{dosovitskiy2020image}.
PatchGame introduces a referential games where two unsupervised models develop a mutual representation through goal-oriented communication~\cite{NEURIPS2021_dac32839}.
SEER~\cite{goyal2022vision} demonstrates the success of strong self-supervised pre-training methods at the scale of billions of images.
With all the recent advances, the latest self-supervised methods have leveraged transformers and iteratively improved upon contrastive and clustering objectives to surpass supervised methods on many key baselines~\cite{pang2022unsupervised,oquab2023dinov2,zhou2022mugs,li2022efficient}.

\paragraph{Unified Models.} 
Other methods leverage the unsupervised nature of GANs to learn good image representations~\cite{donahue2016adversarial,dumoulin2016adversarially,chen2016infogan,nie2020semi}.
BiGAN~\cite{donahue2016adversarial} does joint Encoder-Generator training with a discriminator which jointly discriminates image-latent pair. 
ALI~\cite{dumoulin2016adversarially} uses reparameterized sampling from the encoder output.   
BigBiGAN~\cite{donahue2019large} is the most popular among these methods -- it is a BiGAN with a BigGAN~\cite{brock2018large} generator and a discriminator with additional unary loss terms for image and latent codes. 
In spite of their promising performance for downstream classification tasks, subsequent contrastive pre-training methods that train more quickly, reliably, and with fewer parameters have beaten their performance.

Distinct from GANs, autoencoders are a natural fit for the unified paradigm.
ALAE attempts to learn an encoder-generator map to perform both generation and classification~\cite{pidhorskyi2020adversarial}.
PatchVAE improves on the classification performance of VAE~\cite{kingma2022autoencoding} by encouraging the model to learn good mid-level patch representations~\cite{Gupta_2020_CVPR}.
MAE~\cite{DBLP:journals/corr/abs-2111-06377} and iBOT~\cite{zhou2022ibot} train an autoencoder via masked image modeling, and several other transformer-based methods have been built under that paradigm~\cite{assran2022masked,bao2022beit,huang2022contrastive}.
MAGE~\cite{li2022mage}, which uses a variable masking ratio to optimize for both recognition and generation, is the first method to achieve both high-quality unconditional image generation and good classification results.

\section{Approach}
\label{sec:approach}

\subsection{Diffusion Models Fundamentals}
\label{subsec:diffusion-fundamentals}
Diffusion models first define a forward noising process where gradual Gaussian noise is iteratively added to an image $x_0$, which is sampled from the data distribution $q(x_0)$, to get a completely noised image $x_T$ in $T$ steps. 
This forward process is defined as a Markov chain with latents $x_1,x_2 \dots, x_t, \dots,x_{T-1},x_T$ which represent noised images of various degrees. Formally, the forward diffusion process is defined as\vspace{-0.1in}
\begin{equation}
\label{eq:forward}
\begin{split}
q(x_1,\dots x_T|x_0) &:= \prod_{t=1}^{T}q(x_t|x_{t-1}) \\
q(x_t|x_{t-1}) &:= \mathcal{N} (x_t; \sqrt{1-\beta_t}x_{t-1}, \beta_t\textbf{I})
\end{split}
\end{equation}
where ${\{\beta_t\}}_{t=1}^T$ is the variance schedule and $\mathcal{N}$ is a normal distribution. 
As $T\rightarrow\infty$, $x_T$ nearly is equivalent to the isotropic Gaussian distribution. With $\alpha_t:=1-\beta_t$ and $\Bar{\alpha}_t:=\prod_{i=0}^t\alpha_i$ one can sample a noised image $x_t$ at diffusion step $t$ directly from a real image $x_0$ using
\begin{equation}
\label{eq:make_noisy}
\begin{split}
x_t = \sqrt{\Bar{\alpha}_t}x_0 +\sqrt{1-\Bar{\alpha}_t}\epsilon, \epsilon \sim \mathcal{N}(0, \textbf{I})
\end{split}
\end{equation}
The reverse diffusion process aims to reverse the forward process and sample from the posterior distribution $q(x_{t-1}|x_{t})$ which depends on the entire data distribution.  
Doing this iteratively can denoise a completely noisy image $x_T$, such that one can sample from the data distribution $q(x_0)$. 
This is typically approximated using a neural network $\epsilon_\theta$ as   
\begin{equation}
\label{eq:reverse}
\begin{split}
p_{\theta}(x_{t-1}|x_{t}) := \mathcal{N}\left(x_{t-1}; \frac{1}{\sqrt{\alpha_t}}\left(x_t-\frac{\beta_t}{\sqrt{1-\Bar{\alpha}_t}}\epsilon_\theta(x_t,t)\right),\Sigma_\theta(x_t,t)\right) \\
\end{split}
\end{equation}
When $p$ and $q$ are interpreted as a VAE, a simplified version of the variational lower bound objective turns out to be just a mean squared error loss~\cite{ho2020denoising}. This can be used to train $\epsilon_\theta$ which learns to approximate the Gaussian noise $\epsilon$ added to the real image $x_0$ in Eq. \ref{eq:make_noisy} as
\begin{equation}
\label{eq:loss}
\begin{split}
\mathcal{L}_{\text{simple}}=\mathbb{E}_{x_0,t,\epsilon}[\|\epsilon_\theta(x_t, t) - \epsilon \|_2^2]
\end{split}
\end{equation}
As for $\Sigma_\theta(x_t,t)$, previous works keep it either fixed~\cite{ho2020denoising} or learn it using the original variational lower-bound objective~\cite{nichol2021improved,dhariwal2021diffusion}. 

\vspace{-0.3cm}
\subsection{Diffusion Models Feature Extraction}
\label{subsec:diffusion-feature-extraction}

In this work, we use the guided diffusion (GD) implementation, which uses a U-Net-style architecture with residual blocks for $\epsilon_\theta$. 
This implementation improves over the original \cite{ho2020denoising} architecture by adding multi-head self-attention at multiple resolutions, scale-shift norm, and using BigGAN~\cite{brock2018large}  residual blocks for upsampling and downsampling. 
We consider each of these residual blocks, residual+attention blocks, and downsampling/upsampling residual blocks as individual blocks and number them as $b \in \{1, 2, ..., 37\}$ for the pre-trained unconditional $256{\times}256$ guided diffusion model.

\begin{figure}
    \centering
    \includegraphics[width=\textwidth]{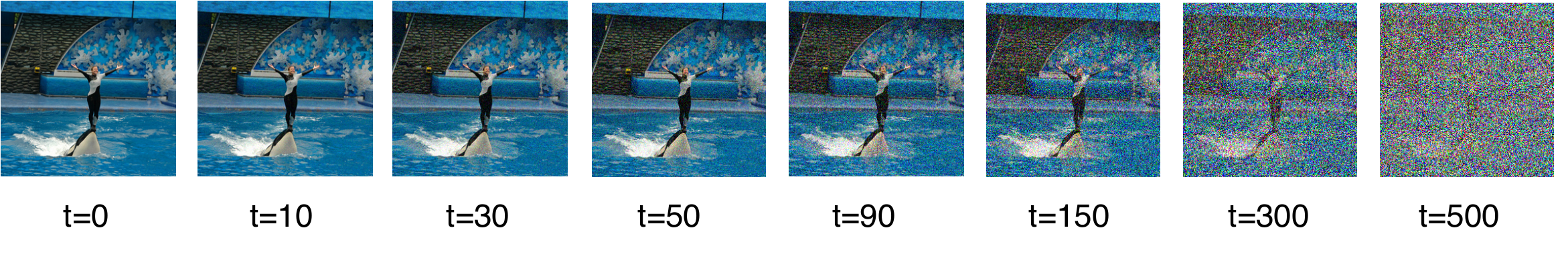}
    \vspace{-0.8cm}
    \caption{Images at different time steps of the diffusion process, with noise added successively. We observe that the best accuracies are obtained at $t = 90$.}
    \label{fig:noise_steps}
\end{figure}

Our feature extraction is parameterized with the diffusion step $t$ and model block number $b$. 
We show an illustration of how input images vary at different time steps in Figure~\ref{fig:noise_steps}.
For feature extraction of image $x_0$, we use Eq. \ref{eq:make_noisy} to get noised image $x_t$. In the forward pass through the network $\epsilon_\theta(x_t,t)$, we use the activation after the block number $b$ as our feature vector $f_\theta(x_0,t,b)$.

\subsection{Linear Probing and Alternatives}

The two most common methods for evaluating the effectiveness of self-supervised pre-training are linear probing and finetuning, and we match the popular recipes documented by VISSL~\cite{goyal2021vissl} to the extent possible.
While correlated, these test different properties of the pre-training.
Linear probing, which learns a batch normalization and linear layer on top of frozen features, tests the utility of the learned feature representations -- it shows whether the pre-training learns disentangled representations, and whether these feature meaningful semantic correlations.
Finetuning, on the other hand, learns a batch normalization and linear layer but with no frozen features.
In the finetuning regime, we treat the pre-training method as an expensive weight initialization method, and retrain the entire architecture for classification.

In this paper, we focus more on the representative capacity of the frozen features, which is of particular interest in areas like fine-grained classification and few shot learning, where data may be insufficient for finetuning.
Additionally, this allows us to make statements with respect to the utility of the learned features, rather than the learned weights. 
We note that the diffusion models are like regular convolutional nets in the sense that they do not natively produce a linear feature, instead generating a series of feature maps at various points in the network.
Thus, similar to other CNNs, we use a combination of pooling and flattening to yield a vector feature representation for each image.

The channel depth and feature map size are naturally quite large, so in addition to standard pooling, we also try other methods.
We investigate multi-layer perceptron heads.
Due to the large size, we also try CNNs as a learned pooling mechanism, and give more complete details for the design in the appendix.
We also investigate the ability of attention heads to perform appropriate aggregation of both spatial and channel information, with full details in the appendix.

\section{Experiments}
\label{sec:experiments}

We first provide some preliminaries for setup and replication purposes, specifically with respect to model architecture, critical hyperparameters, and hardware details.
Then, we give statistics for the datasets we use for our experiments.
We give our primary results in Section~\ref{subsec:main_results} -- we compare our diffusion extraction to baselines as well as competing unified representation methods. %
We provide ablations in Section~\ref{subsec:ablations} to discover optimal block numbers, pooling sizes, and time steps for feature extraction. 
We evaluate the fitness of diffusion for downstream classification tasks by providing results for popular FGVC datasets in Section~\ref{subsec:fgvc}. 
We perform the analysis of our representations in Section~\ref{subsec:analysis} to compare representations both internally (between blocks of the U-Net) as well as externally (between different U-Nets and with other self-supervised learning architectures).

\begin{table*}
\begin{minipage}{0.49\linewidth}
\begin{center}
\caption{
Main results. We compare unified learners in terms of classification and generation at resolution 256. 
}
\label{tab:baselines}
\vspace{0.25em}
\resizebox{.85\textwidth}{!}{
\setlength{\tabcolsep}{8pt}
\begin{tabular}{@{}l|cc@{}}
\toprule
Method & Accuracy & FID  \\
\midrule

BigBiGAN* &   60.8\%	& 28.54 \\
MAGE  & \textbf{78.9\%}  & \textbf{9.10} \\
\midrule
U-Net Encoder	& 64.32\% & n/a \\
\midrule
GD (L, pool $1{\times}1$)  & 61.95\%	& 26.21 \\
GD (L, pool $2{\times}2$)  & 64.96\%	& 26.21 \\
GD (Attention)  & 71.89\% & 26.21 \\
\bottomrule
\end{tabular}
}
\vspace{0.01in}

\hspace{-.02in}\footnotesize{*BigBiGAN's best FID is at generator resolution 128.}%
\end{center}
\end{minipage} 
\hfill
\begin{minipage}{0.49\linewidth}
\begin{center}
\caption{Finetuning results. Non-GD methods use ViT-L. Except for MAGE, all other methods use $224{\times}224$ images.}
\label{tab:full_finetune}
\vspace{0.25em}
    \resizebox{.85\textwidth}{!}{
    \setlength{\tabcolsep}{10pt}
    \begin{tabular}{@{}l c c c@{}} 
        \toprule
        Method & Accuracy \\
        \midrule
        Supervised & 82.5\% \\
        MoCo v3 & 84.1\% \\
        MAE & \textbf{84.9\%} \\
        MAGE & 84.3\% \\
        GD (Linear, pool $2{\times}2$) & 73.17\% \\
        GD (Linear, pool $4{\times}4$) & 73.50\%\\
        \bottomrule
    \end{tabular}
}
\end{center}
\end{minipage} 
\end{table*}

\noindent\textbf{Experiment Details.} Unless otherwise specified, we use the unconditional ADM U-Net architecture from Guided Diffusion~\cite{dhariwal2021diffusion}  with total timesteps $T=1000$.
We use the $256{\times}256$ checkpoint; thus we resize all inputs to this size and use center-crop and flipping for data augmentation.
We use an adaptive average pool to reduce the spatial dimension, followed by a single linear layer.
For linear probing, we train only this single layer. 
We use cross entropy loss with an Adam optimizer~\cite{kingma2017adam}.
We follow the VISSL protocol for linear probing -- 28 epochs, with StepLR at $0.1$ gamma every 8 epochs.
However, we do not use random cropping or batch norm.
For hardware, the majority of our experiments are run on 4 NVIDIA RTX A5000 GPUs.

\noindent\textbf{Datasets.}
The dataset we use for our main result is ImageNet-1k~\cite{deng2009imagenet}.
Additionally, we run ablations and similar explorations on ImageNet-50, which is a selection of 50 classes of ImageNet as also used in \cite{vangansbeke2020scan}.
Please see Table~\ref{tab:datasets} for exact datasets and details.

\subsection{Main Results: ImageNet Classification}
\label{subsec:main_results}
First, we show the promising linear probing performance of diffusion in Table~\ref{tab:baselines}, using settings we select via the ablations described in Section~\ref{subsec:ablations}.
As a baseline, we compare to the diffusion pre-trained classifier, since it uses the same U-Net encoder.
We also offer a comparison to other unified models: BigBiGAN~\cite{donahue2019large} and MAGE~\cite{li2022mage}. 
We outperform BigBiGAN in terms of both generation and classification, especially when BigBiGAN is forced to handle the higher resolution, $256{\times}256$ images.
Hence, diffusion models beat GANs for image classification (and generation).
We acknowledge that diffusion is not yet state-of-the-art compared to classification-only models, with a gap of over $10\%$ top-1 accuracy, or compared to the powerful unified MAGE model.
However, we note that we are unable to completely match the resources necessary to mimic the linear probe settings of other methods.
MAE~\cite{DBLP:journals/corr/abs-2111-06377}, for example, trains their linear layer for 100 epochs with 16,384 images per batch.
Thus, it is difficult to present ``fair'' comparisons with such methods.

We perform finetuning, under similar conditions.
Shown in Table~\ref{tab:full_finetune}, guided diffusion lags behind other methods which use classification specific adjustments.
Regardless, this is a better result than the U-Net encoder by a fair margin (\textbf{+$9.38\%$}), which suggests that guided diffusion is a useful pre-training for classification.

\begin{table*}
\begin{minipage}{0.49\linewidth}
\caption{Attention head ImageNet-1k classification results.}
\label{tab:transformer-inet1k}
\vspace{0.25em}
    \resizebox{1.0\textwidth}{!}{
    \setlength{\tabcolsep}{10pt}
    \begin{tabular}{c  c | c c} 
        \toprule
        $b$ & $t$ & Accuracy (L) & Accuracy (A) \\
        \midrule
        19  & 90    &   55.09\%    &   66.03\% \\
        19  & 150   &   54.77\%    &   64.85\% \\
        24  & 90    &   61.95\%    &   \textbf{71.89\%} \\
        24  & 150   &   61.86\%    &   70.98\% \\
        \bottomrule
    \end{tabular}
}
\end{minipage}  
\hfill
\begin{minipage}{0.49\linewidth}
\caption{Stable Diffusion linear probe results.}%
\label{tab:stable-diffusion}
\vspace{0.25em}
    \resizebox{.98\textwidth}{!}{
    \setlength{\tabcolsep}{10pt}
    \begin{tabular}{@{}l c c | c c@{}} 
        \toprule
        Condition & $b$ & Size & Accuracy \\
        \midrule
        Null Text  & 18 & 512  & 64.67\% \\
        Null Text  & 15 & 512 & 55.77\% \\
        Null Text  & 18 & 256  & 41.37\%\\
        Learnable  & 18 & 512    & \textbf{65.18\%} \\
        \midrule
        Guided Diffusion & 24 & 256 & \textbf{61.86 \%} \\
        \bottomrule
    \end{tabular}
}
\end{minipage} 
\hfill

\end{table*}

As described previously, we also propose several approaches to deal with the large spatial and channel dimensions of U-Net representations.
Naively, we can use a single linear layer with different preliminary pooling, and we show results for various pooling dimensions.
Alternatively, we can use a more powerful MLP, CNN, or attention head to address varying aspects of the feature map height, width, and depth.
For fairness, we train CNNs, MLPs, and attention heads with comparable parameter counts to our linear layers under the various pooling settings. 
We show results for such heads, on ImageNet-50, in Figure~\ref{fig:teaser_fig} (right), with full numerical results and model details in the appendix.
We note that the attention head performs the best by a fair margin.
In Table~\ref{tab:transformer-inet1k}, we try the best-performing attention head on ImageNet (all classes), and find it significantly outperforms the simple linear probe, regardless of pooling.
This suggests the classification head is an important mechanism for extracting useful representations from diffusion models, and it could be extended to other generative models.

\begin{figure}
    \centering
    \includegraphics[width=\textwidth]{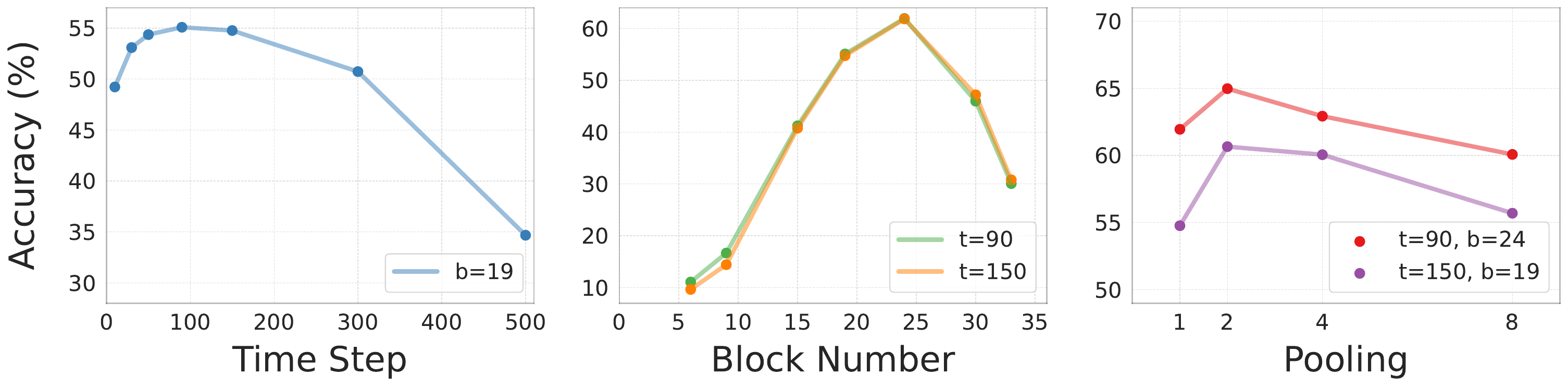}
    \caption{Ablations on ImageNet (1000 classes) with varying block numbers, time steps, and pooling size, for a linear classification head on frozen features. We find the model is least sensitive to pooling, and most sensitive to block number, although there is also a steep drop-off in performance as inputs and predictions become noisier.}
    \label{fig:bn_t_acc}
\end{figure}

\subsubsection{Ablations}
\label{subsec:ablations}
As shown in Figure~\ref{fig:teaser_fig}, extracting good features from diffusion models requires careful consideration of noise step, block number, and pooling size.
We initiate a search of that hyperparameter space for ImageNet.
We set a search space of roughly log-equidistant time steps for the noise.
We try several blocks at even intervals around the U-Net bottleneck.
We also address the feature height and width (pooling).
From our linear search, shown in Figure~\ref{fig:bn_t_acc}, we find $t$ should be set to $90$ and $b$ to $24$.
However, as we discuss in Section~\ref{subsec:fgvc}, we find that such settings are at least somewhat data dependent.
Thus, while in this work we distill some general settings and principles, automatic selection and combination of features could be explored in future work.

For further ablations, we explore to what extent our idea is valid for other diffusion models.
We specifically examine stable diffusion, training a classifier on frozen features for 15 epochs, with $t$ fixed at $150$.
Thus, in Table~\ref{tab:stable-diffusion}, we show that stable diffusion features also lend themselves well to classification.
 Critically, this means not only that our approach is flexible, but that lighter diffusion models with better performance that are developed in the future could be immediately leveraged as unified representation models by our method.

\begin{figure}
    \centering
    \includegraphics[width=1.0\linewidth]{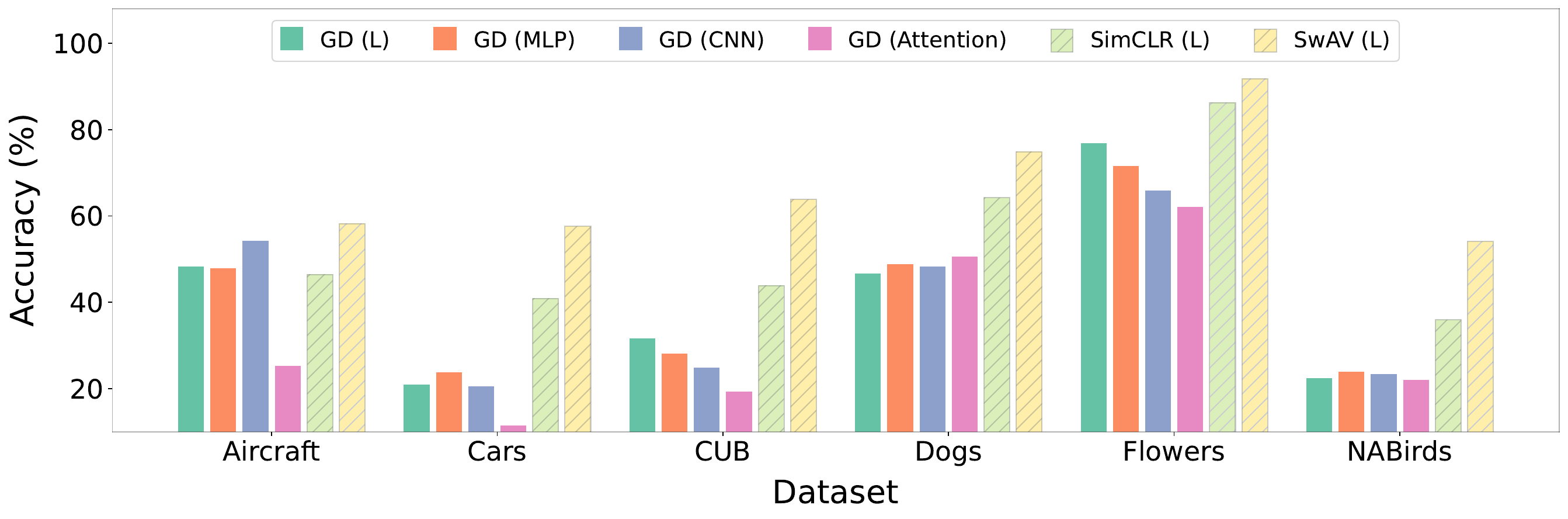}
    \caption{Fine-Grained Visual Classification (FGVC) results. We train our best classification heads from our ImageNet-50 explorations on FGVC datasets (denoted with GD), and compare against the results from linear probing a SimCLR ResNet-50 on the same datasets. Linear is denoted by (L). While SimCLR and SwAV tend to perform better, diffusion achieves promising results, slightly outperforming SimCLR for Aircraft.}
    \label{fig:fgvc}
\end{figure}

\begin{figure}
    \centering
    \includegraphics[width=1.0\linewidth]
    {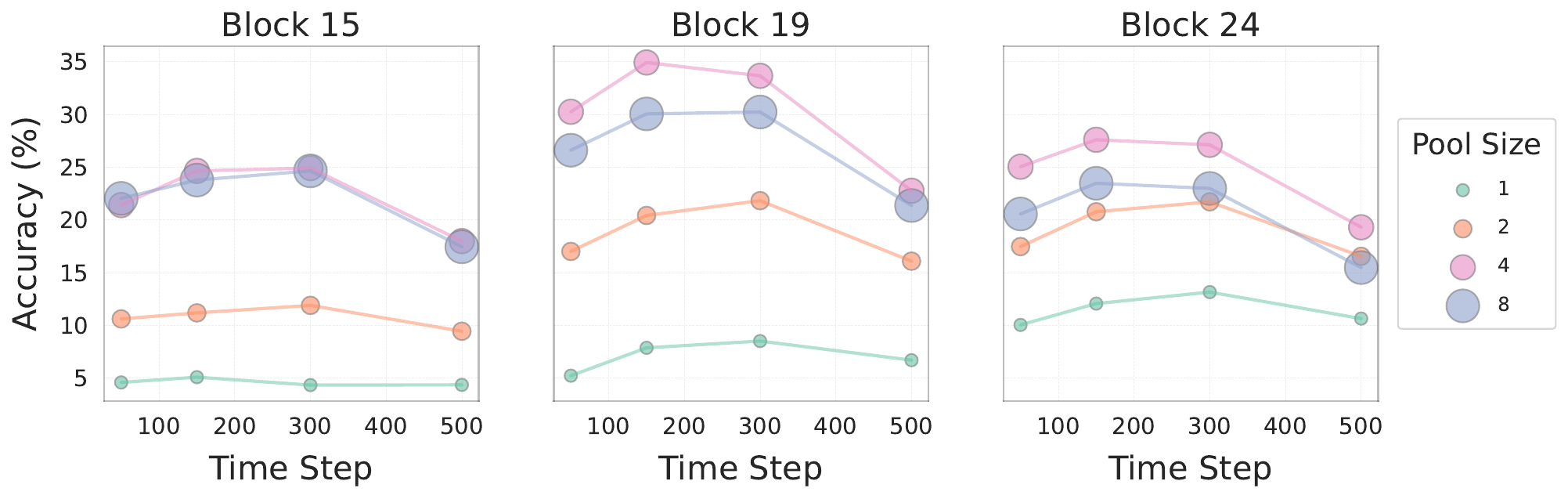}
    \caption{FGVC feature extraction analysis. We show accuracy for different block numbers, time steps, and pooling sizes. Block 19 is superior for FGVC, in contrast to ImageNet where 24 was ideal.}
    \label{fig:cub_search}
\end{figure}

\subsection{Results: Fine-grained Visual Classification (FGVC)}
\label{subsec:fgvc}
Here, we give results for applying our method in the transfer setting to the datasets defined in Table~\ref{tab:datasets}.
We use both standard linear probing, as well as each of our classification heads (with their best ImageNet-50 configurations). 
We show these results in Figure~\ref{fig:fgvc}.
Note that there is a performance gap between the diffusion model and SimCLR, regardless of classification head used.
One notable exception is Aircraft, where diffusion outperforms SimCLR for 3 of the 4 heads; this is indicative of its promising performance.

Additionally, we find that feature selection is not trivial, and often the settings that work for various FGVC datasets do not correspond to the ideal ImageNet settings.
For example, consider that attention, the best head for ImageNet-50, tends to perform the worst for FGVC.
This may be due to their reliance on the amount of data to learn properly.
Furthermore, as we explore the feature selection problem on CUB on Figure~\ref{fig:cub_search}, we find that the ideal block number for ImageNet ($b=24$) underperforms substantially for CUB compared to $b=19$.
Hyperparameter changes that have a more subdued effect on ImageNet, such as pooling size, can result in up to $3\times$ change in performance on accuracy for CUB.
Thus, determining a more robust feature selection procedure or introducing some regularization during the diffusion training might be important future work to make transfer more reliable.

\begin{figure}
    \centering
    \includegraphics[clip,width=0.9\textwidth]
    {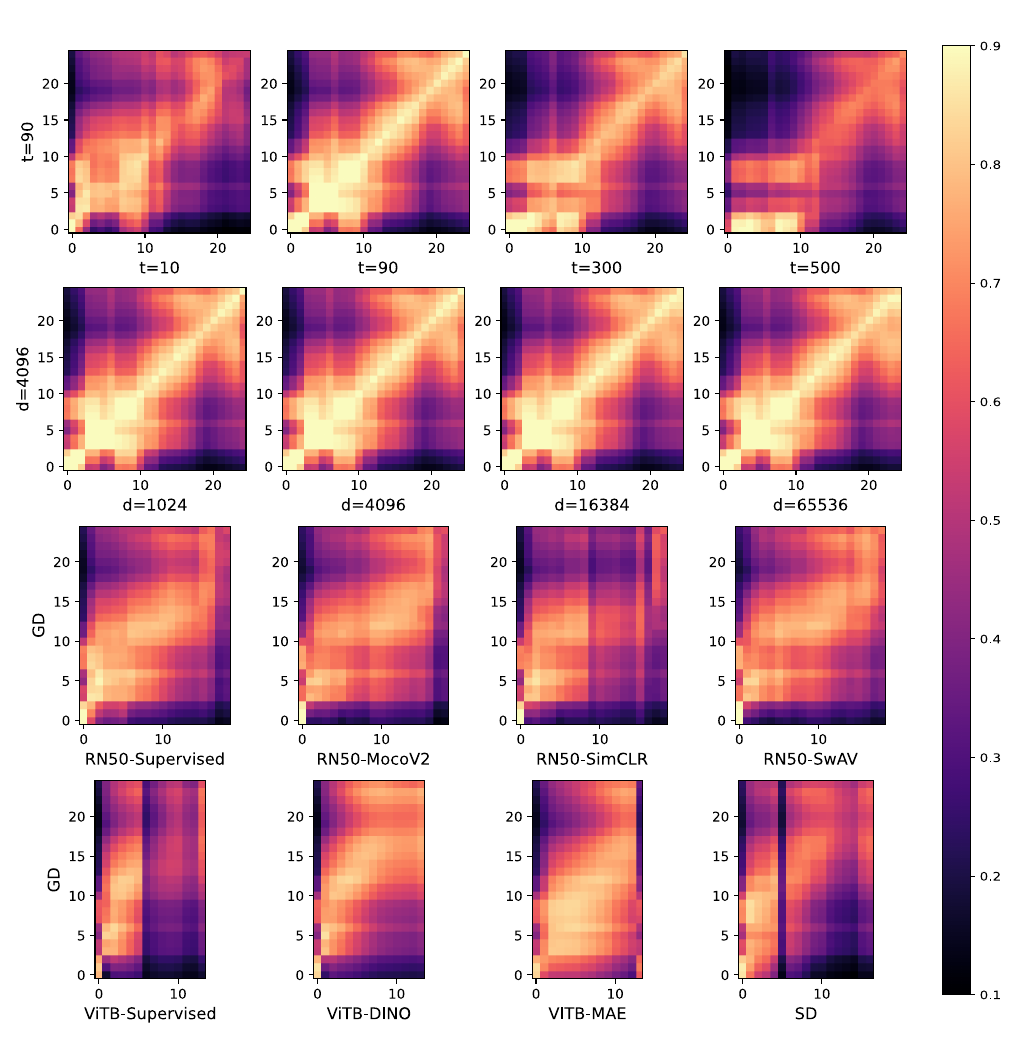 }
    \caption{Feature representation comparisons via centered kernel alignment (CKA). On the top 2 rows, we compare guided diffusion (GD) representations between its own layers, at varying time steps and feature size. On the bottom 2 rows, we compare GD, with standard $t=90$ and $d=4096$, against both ResNets and ViTs with various pre-training methods. For the bottom right corner we compare against Stable Diffusion (SD), $b=18, \text{size}=512$.}
    \label{fig:cka-expanded}
\end{figure}

\subsection{Representation Analysis}
\label{subsec:analysis}
We use linear centered kernel alignment (CKA)~\cite{pmlr-v97-kornblith19a} to find the degree of similarity between the representations of different blocks of the diffusion model.
Following conventions from prior work that use samples for CKA~\cite{Gwilliam_2022_CVPR,walmer2023teaching}, we use the 2,500 image test set of ImageNet-50 (see Table~\ref{tab:datasets}).
We first examine differences in the representations between guided diffusion blocks at various time steps and feature dimensions (pooling size) within our diffusion method in Figure~\ref{fig:cka-expanded}.
We also compare our standard setting ($t=90$ and $d=4096$) against ResNet-50 and ViT representations with a representative set of popular pre-training methods, as well as stable diffusion. 
For ResNet-50, we extract the features from each bottleneck block while for ViT we extract features from each hidden layer.

We note that the early layers tend to have higher similarity in all cases, suggesting that diffusion models likely capture similar low-level details in the first few blocks.
Also note the impact of the time step: the representations are very dissimilar at later layers when the representations are computed using images from different noise time steps.
However, interestingly, we find that around the bottleneck, the layers of GD tend to have similar representations to ResNets and ViTs, suggesting that GD's later layers naturally learn discriminative properties.
This further supports our findings in Table~\ref{tab:baselines} and Table~\ref{tab:transformer-inet1k}, where we show the promising classification performance with GD features.

\section{Conclusion}

In this paper, we present an approach for using the representations learned by diffusion models for classification tasks.
This re-positions diffusion models as potential state-of-the-art unified self-supervised representation learners. 
We explain best practices for identifying these representations and provide initial guidance for extracting high-utility discriminative embeddings from the diffusion process.
We demonstrate promising transfer learning properties and investigate how different datasets require different approaches to feature extraction.
We compare the diffusion representations in terms of CKA, both to show what diffusion models learn at different layers as well as how diffusion representations compare to those from other methods.

\noindent\textbf{Broader Impacts.}
With our paper, we analyze algorithms; we do not provide new real-world applications.
Nevertheless, our work deals with image generation, which carries ethical concerns with respect to potential misinformation generation.
However, we do not improve over existing generation approaches, so the potential harms seem negligible.

\noindent\textbf{Limitations.}
Training diffusion models, even just for linear probing, is very computationally intensive.
So, we could not provide an analysis of variability in this work.
Nevertheless, our work is an important first step for leveraging the capacity of diffusion models for discriminative tasks.

\clearpage
{\small
\bibliographystyle{ieeetr}
\bibliography{main}
}

\clearpage
\appendix

\setcounter{page}{1}

{
\centering
\Large
\textbf{Diffusion Models Beat GANs on Image Classification} \\ 
     {\small Supplementary Material} \\
\vspace{2.0em}
} 
\section{Method Details}

\noindent\textbf{Convolutional Heads.}
In the context of standard linear probing, we treat our convolutional head as a replacement for the feature pooling step.
That is, instead of an adaptive average pool followed by a linear layer, we train a convolutional head, followed by a linear layer.
While we explore different channel dimensions, the architecture consists of 2 blocks, each with a $2{\times}2$ convolution followed by a $2{\times}2$ maximum pooling operation. 
We perform a $2{\times}2$ adaptive average pool, flattening the result so it can be used by a linear layer.
We indicate the convolution heads we train in Table~\ref{tab:cnn-results}, where the first (input) channels is always 1024 (the number of channels of the feature maps at the chosen U-Net block), but the output channels of both learned convolution operations is treated as a hyperparameter.

\noindent\textbf{Attention Heads.} 

In addition to applying convolutional heads, we experiment with a more sophisticated architecture applying Transformer blocks. Specifically, we first use a $2{\times}2$ pooling layer to reduce the spatial resolution of the feature maps to $8{\times}8$. Each token has a channel dimension of $1024$, corresponding to the number of channels of the feature maps extracted from the U-Net block. We then flatten these features to generate $64$ tokens for our Transformer. We append a CLS token to these set of tokens which we then use to make the final classification. We follow the standard Transformer block architecture consisting of Layer Normalization and QKV-Attenton. We treat the number of attention blocks as a hyperparameter and our choices and the corresponding results are shown in Table~\ref{tab:attention-results}. 

\section{Experiment Details}

We provide additional clarifying details for our experiments.
First, the ablations shown in Figure~\ref{fig:bn_t_acc} are the result of training on ImageNet for 15 epochs, instead of our complete linear probing recipe, due to resource constraints. 
Second, the stable diffusion model we compare to was text-guided during its pre-training, and thus, unlike our guided diffusion checkpoint, it was not fully unsupervised.
Finally, we provide parameter count comparisons for our method and the other unified representation learners in Table~\ref{tab:parameter-counts}.

\section{Ablations}

\begin{figure}[h]
    \centering
    \includegraphics[width=1.0\linewidth]{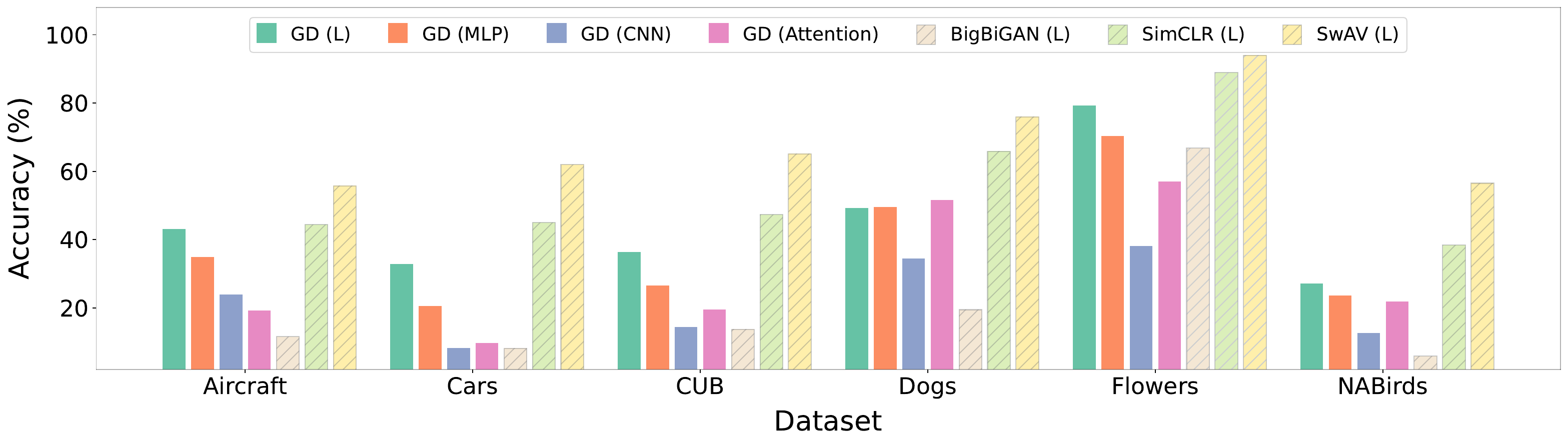}
    \caption{Fine-Grained Visual Classification (FGVC) results, where the classification heads were trained with random resize cropping. Similar to Figure~\ref{fig:fgvc} in terms of methods chosen, except now we provide results for BigBiGAN also.}
    \label{fig:fgvc-supp}
\end{figure}

As noted, in the main paper, we use only random flipping for our linear probe, unlike standard settings which also use a random resized crop.
This is in part due to the fact that while standard unsupervised pre-training would use random cropping, guided diffusion does not.
However, for the sake of thorough comparison, we also try FGVC linear probing with the standard, random resize cropping setup.
We provide these results in Figure~\ref{fig:fgvc-supp}.
Note that the diffusion heads tend to perform worse with the extra augmentations, possibly due to a lack of learned invariance.
We also include BigBiGAN.
Note that this model was implemented in Tensorflow, and we also perform the linear probing in Tensorflow for this model only.

\begin{table*}[h]
\begin{minipage}{0.5\linewidth} %
\caption{Dataset details used in this work.}
\label{tab:datasets}
\vspace{0.25em}
    \resizebox{.98\textwidth}{!}{
    \setlength{\tabcolsep}{10pt}
    \begin{tabular}{@{}l c c c@{}} 
        \toprule
        Dataset & \multicolumn{1}{c}{\#Cls} & \multicolumn{1}{c}{\#Train} & \multicolumn{1}{c}{\#Test} \\
        \midrule
        Aircraft \cite{maji13fine-grained} & 100 & 6,667 & 3,333 \\
        Cars \cite{KrauseStarkDengFei-Fei_3DRR2013} & 196 & 8,144 & 8,041 \\
        CUB \cite{WahCUB_200_2011} & 200 & 5,994 & 5,794 \\
        Dogs \cite{KhoslaYaoJayadevaprakashFeiFei_FGVC2011} & 120 & 12,000 & 8,580 \\
        Flowers \cite{Nilsback08} & 102 & 2,040 & 6,149 \\
        NABirds \cite{7298658} & 555 & 23,929 & 24,633 \\
        ImageNet \cite{deng2009imagenet} & 1000 & 1.3mil & 50,000 \\
        ImageNet-50 \cite{vangansbeke2020scan} & 50 & 64,274 & 2,500 \\
        \bottomrule
    \end{tabular}
}
\end{minipage} %
\hfill
\begin{minipage}{0.5\linewidth}
\caption{Linear probing result with our best setting and augmentation.}
\label{tab:linear-aug-result}
\vspace{0.25em}
    \resizebox{.98\textwidth}{!}{
    \setlength{\tabcolsep}{8pt}
\begin{tabular}{@{}l|c@{}}
\toprule
Setting & Accuracy  \\
\midrule
Linear Probe ($t=90,b=24$, Pool=2) &   \textbf{65.07 \%}\\

\bottomrule
\end{tabular}
}
\end{minipage}

\end{table*}

We also try data augmentations for our best linear probe setting, with results provided in Table~\ref{tab:linear-aug-result} after 15 epochs. We do not observe any significant improvement over the original setting without augmentation.

\section{Figure~\ref{fig:teaser_fig} Details}

Now, we provide more details, in Table~\ref{tab:lin-mlp-results}, Table~\ref{tab:cnn-results}, and Table~\ref{tab:attention-results} for the information shown in Figure~\ref{fig:teaser_fig}.
We show exact accuracies and parameter counts.
We also specify our hyperparameter selection for each head.

\begin{table*}[h]
\begin{minipage}{0.475\linewidth}
\caption{Linear and MLP results. For linear, 1k, 4k, 16k, and 65k indicate the size of the feature after pooling and flattening. For MLP, the first number is the size of the feature after pooling and flattening, and the succeeding numbers are hidden sizes of layers before the last (classification) layer.}
\label{tab:lin-mlp-results}
\vspace{0.25em}
    \resizebox{.98\textwidth}{!}{
    \setlength{\tabcolsep}{10pt}
    \begin{tabular}{@{}l | c c@{}} 
        \toprule
        Head & Params & Accuracy \\
        \midrule
        Linear-1k    & 1M & 87.20\% \\
        Linear-4k    & 4M & 89.41\% \\
        Linear-16k   & 16M & 89.58\% \\
        Linear-65k   & 65M & 88.28\% \\
        MLP-1k-2k    & 4M & 87.93\% \\
        MLP-4k-2k    & 10M & 88.50\% \\
        MLP-4k-2k-2k & 14M & 88.67\% \\
        MLP-16k-2k   & 34M & 89.76\% \\
        \bottomrule
    \end{tabular}
}
\end{minipage} 
\hfill
\begin{minipage}{0.475\linewidth}
\caption{CNN head results. Channel sizes are separated by dashes. The first is the channel size of the input feature maps, the next is the output channels from the first convolution, and the last is the output dimension of the second convolution. These are treated as hyperparameters. For more detail, see our code.}
\label{tab:cnn-results}
\vspace{0.25em}
    \resizebox{.98\textwidth}{!}{
    \setlength{\tabcolsep}{8pt}
    \begin{tabular}{@{}l|ccc@{}}
        \toprule
        Head & Params & Accuracy \\
        \midrule
        CNN-1k-256-256    & 2.5M & 88.98\% \\
        CNN-1k-512-256    & 3.5M & 88.67\% \\
        CNN-1k-1k-256   & 6M & 88.93\% \\
        CNN-1k-1k-1k   & 12M & 89.50\% \\
        CNN-1k-2k-512    & 14M & 89.19\% \\
        CNN-1k-2k-2k     & 32M & 89.76\% \\
        CNN-1k-4k-2k  & 48M & 89.89\% \\
        CNN-1k-4k-2.5k   & 66M & 89.50\% \\
        \bottomrule
\end{tabular}
}
\end{minipage} 
\end{table*}

\begin{table*}[h]
\begin{minipage}{0.6\linewidth}
\begin{center}
\caption{Attention head results. The hyperparameters are denoted following the dashes. The first hyperparameter is similarly the channel size of the input feature maps and the next is the number of Transformer blocks used.}
\label{tab:attention-results}
\vspace{0.25em}
    \resizebox{.7\textwidth}{!}{
    \setlength{\tabcolsep}{10pt}
    \begin{tabular}{@{}l | c c@{}} 
        \toprule
        Head & Params & Accuracy \\
        \midrule
        Attention-1K-1 & 13.7M & 91.67\% \\
        Attention-1K-2 & 26.2M & 91.58\% \\
        Attention-1K-3 & 38.8M & 92.01\% \\
        Attention-1K-4 & 51.4M & 92.27\% \\
        Attention-1K-5 & 64.0M &  92.71\% \\
        \bottomrule
    \end{tabular}
}
\end{center}
\end{minipage} 
\hfill
\begin{minipage}{0.35\linewidth}
\begin{center}
\caption{Parameter counts of major unified unsupervised representation learning methods. For each, we consider the whole system, not just the encoding network.}
\label{tab:parameter-counts}
\vspace{0.25em}
    \resizebox{.7\textwidth}{!}{
    \setlength{\tabcolsep}{8pt}
\begin{tabular}{@{}l|c@{}}
\toprule
Method & \# Params  \\
\midrule
BigBiGAN & 502M  \\
MAGE & 439M \\
Ours & 553M \\
\bottomrule
\end{tabular}
}
\end{center}
\end{minipage}
\end{table*}

\end{document}